\documentclass[10pt]{article}
\usepackage[legalpaper, margin=1in]{geometry}

\usepackage{amssymb}
\newcommand\Tau{\mathcal{T}}

\usepackage{times}
\usepackage{epsfig}
\usepackage{graphicx}
\usepackage{amsmath}
\usepackage{amssymb}
\usepackage{tcolorbox}
\usepackage{caption}
\usepackage{subcaption}

\usepackage{url}
\usepackage{booktabs} 
\usepackage{nicefrac} 
\usepackage{microtype} 
\usepackage{xcolor} 
\usepackage{enumitem}
\usepackage[ruled,vlined]{algorithm2e}
\usepackage{caption}
\usepackage{multirow}

\usepackage{xr}

\usepackage{authblk}

\begin{document}

\title{Investigating the Impact of Weight Sharing Decisions on Knowledge Transfer in Continual Learning} 
\author[1,*]{Josh Andle}
\author[2]{Ali Payani}
\author[1]{Salimeh Yasaei Sekeh}
\affil[1]{University of Maine}
\affil[2]{Cisco Systems Inc.}
\affil[*]{Corresponding Author: joshua.andle@maine.edu}
\setcounter{Maxaffil}{0}
\renewcommand\Affilfont{\itshape\small}

\date{}

\maketitle

\begin{abstract}
Continual Learning (CL) has generated attention as a method of avoiding Catastrophic Forgetting (CF) in the sequential training of neural networks, improving network efficiency and adaptability to different tasks. Additionally, CL serves as an ideal setting for studying network behavior and Forward Knowledge Transfer (FKT) between tasks. Pruning methods for CL train subnetworks to handle the sequential tasks which allows us to take a structured approach to investigating FKT. Sharing prior subnetworks’ weights leverages past knowledge for the current task through FKT. Understanding which weights to share is important as sharing all weights can yield sub-optimal accuracy. This paper investigates how different sharing decisions affect the FKT between tasks. Through this lens we demonstrate how task complexity and similarity influence the optimal weight sharing decisions, giving insights into the relationships between tasks and helping inform decision making in similar CL methods. We implement three sequential datasets designed to emphasize variation in task complexity and similarity, reporting results for both ResNet-18 and VGG-16. By sharing in accordance with the decisions supported by our findings, we show that we can improve task accuracy compared to other sharing decisions.

\end{abstract}

\section{Introduction}
Continual Learning (CL) has garnered growing interest in recent years, focused on preventing the catastrophic forgetting (CF) that occurs when a network is retrained on new data. By enabling sequential learning in neural networks, CL has applications in dynamic environments including autonomous navigation and shows promise in areas where a network is expected to draw upon past knowledge to learn new information. The latter case is termed forward knowledge transfer (FKT) and occurs when past information can be reused to benefit the handling of new tasks. The potential for FKT is a benefit of handling sequential training with a single network which remains insufficiently realized. \\
Over the course of investigations on CL, three main types of approaches have been developed: Replay methods, Regularization methods, and Architectural methods. Replay methods typically avoid forgetting by training generative models such as GANs to synthesize samples of data from past tasks for retraining (Shin et al., 2017; Wu et al., 2018). Regularization models impose regularization terms on the updates of network weights to ensure that weights useful for past tasks aren't significantly modified (Kirkpatrick et al., 2017; Zenke et al., 2017). Architectural methods generally modify or use the network's architecture to handle multiple tasks, such as by expanding the network or pruning (Mallya and Lazebnik, 2018; Hung et al., 2019).\\
We focus on pruning-based methods as we can control which subnetworks are connected through weights, allowing us to influence what information a given subnetwork has access to when handling a task by adjusting which filters it can rely on. This allows for a direct investigation into how we can predict if a given subnetwork will be useful in the training of a new task. Intuition may suggest that sharing information or weights from all past subnetworks is optimal for ensuring the best learning and accuracy on a new task, however it has been shown that this is not the case and sharing a subset of past information can outperform this baseline (Wang et al., 2020). While this was shown when sharing individual weights, we show that it remains true even when sharing entire filters, where interference from sharing different weights within a given filter cannot occur.\\
Similar past works have investigated different weight sharing strategies, often based upon the gradient of loss with respect to shareable weights (Wang et al., 2020; Sokar et al., 2021), but insufficient insight has been given as to why certain weights are beneficial for sharing or not. To this end we have put together three different sequences of tasks meant to emphasize differences in task complexity and similarity and analyzed how different task ordering or subnetwork sharing can promote FKT. These experiments are performed on both a VGG-16 and ResNet-18 architecture to investigate how the architecture may influence subnetwork sharing strategies. In this paper, we make the following contributions:
\begin{enumerate}
    \item To improve interpretability we implement a pruning and sharing strategy which ensures that a given filter's feature representation remains consistent in both its original task and any task for which it's shared.
    \vspace{-0.2cm}
    \item We methodically investigate how sharing decisions can be made based on the properties of available subnetworks to improve accuracy on a new task. 
    \vspace{-0.2cm}
    \item We evaluate different sharing strategies leveraging these subnetwork properties on three CL datasets.
\end{enumerate}

\section{Related Works}
\textbf{Learned Weight Sharing}
Over the past several years a handful of papers have been published investigating potential methods for sharing previously trained weights during pruning-based Continual Learning (CL). Initial work on pruning-based methods for CL learn subnetworks for each task but shared all frozen weights for new tasks (Mallya and Lazebnik, 2018). Later works showed the potential of binary masking for enabling shared, fixed weights to contribute differently to multiple tasks (Mallya et al., 2018; Serra et al. 2018). Subsequent work additionally introduced iterative pruning and conditional network expansion (Hung et al., 2019), and began to measure knowledge transfer in this masking setting (Golkar et al., 2019). Notably our experiments use a fixed network capacity as this is a common constraint imposed on benchmark CL methods. \\
Later works build off of the earlier masking techniques by implementing iterative approaches to learn the optimal mask of frozen weights for the current task (Wang et al., 2020; Dehkovich et al., 2023). These works identify masks by evaluating the gradient of loss with respect to the frozen weights. By sharing at the parameter level rather than entire filters, they don't provide the insights into which subnetworks are providing useful information for FKT which we aim to investigate here. Additional investigation has been done in looking for optimal subnetworks from the perspective of the Lottery Ticket Hypothesis (Chen et al., 2020; Kang et al., 2022), however these focus more on the learning of a task mask for sparse training than weight sharing. Lastly, some work has investigated the use of soft or non-binary masking (Konishi et al., 2023) and weight sharing (Sokar et al., 2021) where shared weights may still be trained. By contrast, we focus on the setting in which no shared weights may be updated after they have been frozen.

\textbf{Subnetwork Properties}
When pruning we implement constraints on the resulting structure of weights in the network to maintain filter behavior in the current task and any future task for which they're shared. Previous CL work has provided motivations for similar constraints but focus on preventing CF rather than feature shift during weight sharing (Golkar et al. 2019; Jung et al. 2020). The similarity between tasks within CL has previously been calculated using the difference in loss between an independently trained network and one which shares previously trained weights (Ke et al., 2020). We implement an equivalent definition of task similarity when sharing weights and evaluating FKT. When measuring the usefulness of a given subnetwork we utilize the measure of connectivity as a linear estimator of layer dependencies. Previous work has been done showing that pruning based on connectivity can improve overall network accuracy in CL (Andle and Yasaei Sekeh, 2022).

\begin{figure*}[t]
    \centering
    \includegraphics[width=0.9\textwidth]{"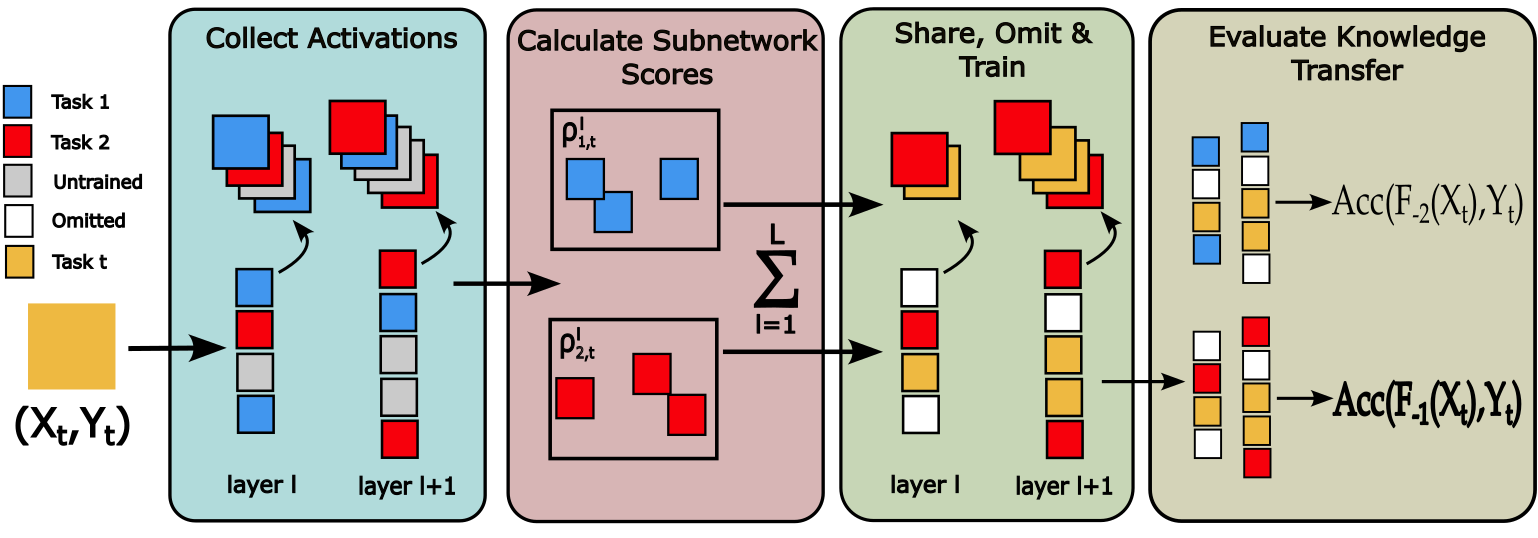"}
    \caption{Given a network $F$ composed of subnetworks $S_{t'<t}$ optimized for different tasks, when learning a new task $\Tau_t$ we want to identify and leverage only the most useful subnetworks for $\Tau_t$. To achieve this we A.) collect the activations over all network layers, B.) compute usefulness scores over only the activations of each individual subnetwork, and C.) temporarily omit any subnetworks deemed non-useful for the current task $\Tau_t$. D.) After training on $\Tau_t$, we compare the resulting accuracy among different sharing decisions. The aim is to identify the combination of available subnetworks for sharing and omission which maximizes the accuracy on $\Tau_t$ after training.}
    \label{Fig:MethodOverview}
\end{figure*}

\section{Methods}
\subsection{Notation}
\begin{itemize}
    \item $F$: The full network composed of L layers $F^1$-$F^L$
    \vspace{-0.2cm}
    \item $f^l_{i}$: A single filter i in layer l
        \vspace{-0.2cm}
    \item $\Tau$: The dataset composed of T sequential 
    tasks $\Tau_1$-$\Tau_T$
        \vspace{-0.2cm}
    \item $S_t$: The subnetwork used for a given task $\Tau_t$
        \vspace{-0.2cm}
    \item $S_{t'<t}$: A subnetwork for task $\Tau_{t'}$ trained before $\Tau_t$
        \vspace{-0.2cm}
    \item $\omega^*_t$: The weights optimized and frozen for a task $\Tau_t$
        \vspace{-0.2cm}
    \item $M_t$: Binary mask with ones at the indices of $S_t$
        \vspace{-0.2cm}
    \item $F_{+t}$ and $F_{-t}$: The network $F$ with subnetwork $S_t$ either shared or omitted, respectively
\end{itemize}

\begin{figure*}[t]
    \begin{subfigure}[b]{\textwidth}
    \centering
        \includegraphics[width=0.95\columnwidth]{"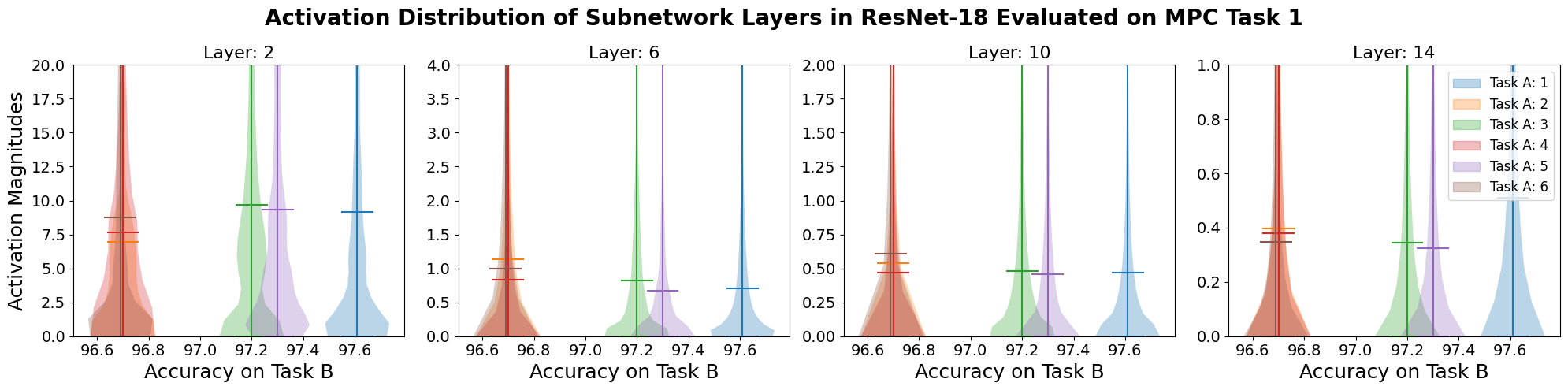"}
    \end{subfigure}
    \begin{subfigure}[b]{\textwidth}
    \centering
        \includegraphics[width=0.95\columnwidth]{"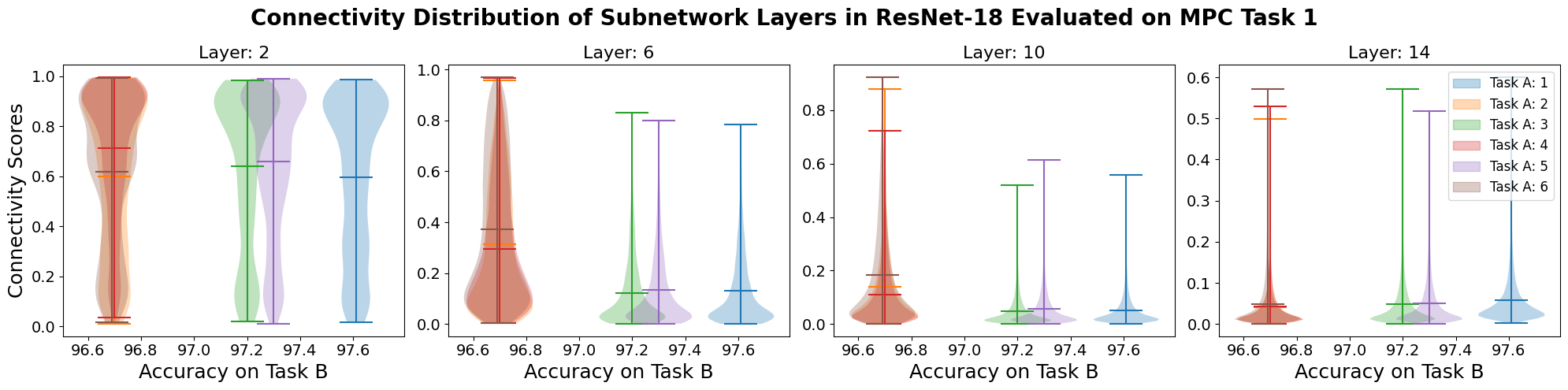"}
    \end{subfigure}
\caption{The distribution of squared activations (top) and connectivities (bottom) values over four layers of ResNet-18 are plotted when Task $\beta$=1. The shaded area reflects the KDE of values for the given layer. The activations often do not differ significantly between tasks $\Tau_\alpha$ despite PMNIST tasks (1,3, and 5) providing greater accuracy than CIFAR tasks (2,4, and 6). By contrast we consistently observe lower mean connectivities in subnetworks which provide greater accuracy.}
\label{Fig:TwoTask}
\end{figure*}

\subsection{Continual Learning}
During the sequential training of a deep neural network (DNN) denoted as $F$, we train on dataset $\Tau$ consisting of tasks $\Tau_1$-$\Tau_T$. Each task is comprised of data $(\mathbf{X}_t, Y_t)$, with $N_t$ samples belonging to classes $C_t$ for a given task $\Tau_t$. $F$ is composed of $L$ layers $F^1$-$F^L$, whose filters are denoted $f^l_i$ for a given layer l. For simplicity, we will use the term filters when referring to either convolutional or linear layers in the network unless additional context is needed. \\
After training finishes for the first task $\Tau_1$, each layer is pruned such that $k\%$ of the unfrozen filters in each layer have their weights set to 0, resulting in a subnetwork $S_1$ of filters deemed useful for the task and a binary mask $M_1$ is stored for all layers in $F$ with elements $0$ corresponding to pruned weights. The element-wise multiplication $M_1 \odot F$ yields the weights which remained after the pruning of task $\Tau_1$. After finetuning, the weights in $S_1$ are frozen such that their gradients are fixed to 0 and they cannot be pruned. The network is then trained on the next task $\Tau_t$. When training on subsequent tasks, each past subnetwork $S_{t'<t}$ can either be omitted, in which case $M_{t'} \odot F \longleftarrow0$, or it is included for the training of task $\Tau_t$. The constraints on included weights are covered in the next subsection. \\
This process is repeated for all tasks in $\Tau$, generating a different mask $M_t$ and corresponding subnetwork $S_t$ for each. As we work with the task-incremental setting of CL, a separate classifier head $H_t$ is trained for each task and used for inference on task $\Tau_t$. For clarity, when referring to a network $F$ in which a single subnetwork $S_t$ is included or omitted, we'll denote it as $F_{+t}$ or $F_{-t}$ respectively.Training on a given task $\Tau_t$ is done using the cross entropy loss function, with loss $\mathcal{L}_t$ calculated as:
\begin{align}\label{Def:Loss}
\mathcal{L}_t := \mathbb{E}_{(\mathbf{X}_t,Y_t)\in \Tau_t}\big\{\ell\big(F_{\omega_t,\omega^*_{t'}}(\mathbf{X}_t), Y_t\big)\big\},
\end{align}
where $\ell$ is a loss function (e.g. cross entropy), $\omega_t$ are the trainable weights, and $\omega^*_{t'}$ are any frozen weights shared from past tasks $\Tau_{t'<t}$ in the network $F$. 
\subsection{Interpretable Pruning}\label{section:interpretablepruning}
Previous methods for pruning-based CL have utilized unstructured pruning to good effect (Wang et al., 2020). Compared to structured pruning where entire filters/channels need to be pruned or frozen, unstructured pruning allows for flexibility in how the sparsity can be distributed within a layer. While this has shown good results, the effect on the network during weight sharing is less interpretable. \\
When a filter $f^l_{i,t'}$ is frozen after training converges on a given task $\Tau_{t'}$, that filter is expected to capture useful features for that data. In the unstructured setting, if $f^l_{i,t'}$ is then shared for a task $\Tau_t$ and either a subset of the weights within it are not also shared or weights in the filters of $F^{l-1}$ upon which it depends aren't shared, then the feature that it represents may no longer reflect its behavior in $\Tau_{t'}$. Similarly, if any of the weights within $f^l_{i,t'}$ or upon which it depends in the prior layers are allowed to be trained for the new task, or if multiple unstructured subnetworks are shared together, then the represented feature will change as well. This doesn't result in CF, as the original task's mask will omit these newly trained weights when revisiting the original task, however, it does change the features represented by the filters and hence the interpretability of \textit{why} $f^l_{i,t'}$ is useful for $\Tau_t$. As our focus is on coming to interpretable conclusions about the properties of useful subnetworks that facilitate FKT, we implement structured pruning throughout the experiments. To ensure that a shared filter reflects the same feature as it did in its original task, multiple constraints have to be enforced on the structure of subnetworks:
\begin{enumerate}
    \item Once filter $f^l_i$ is frozen, its weights cannot be updated.
    \vspace{-0.2cm}
    \item If $f^{l}_i$ is shared, filters in previous layers with which it is connected by non-zero weights are shared as well.
    \vspace{-0.2cm}
    \item During pruning, when a filter $f^l_i$ is pruned, all weights in $F^{l+1}$ connected to $f^l_i$ are pruned. 
\end{enumerate}
The last constraint is less intuitive, but consider the case where multiple subnetworks are shared. A filter $f^l_{i,\alpha}$ frozen during task $\Tau_\alpha$ may have a nonzero weight to a filter $f^{l-1}_{j,\beta}$ which was originally pruned during task $\Tau_\alpha$ but later retrained and frozen for task $\Tau_{\beta}$. This leads to unpredictable behavior of $f^l_{i,\alpha}$ depending on whether or not $S_\beta$ is also being shared. To prevent this behavior, whenever we prune a filter in a given layer $l$ we must also prune all weights in layer $l+1$ which connect to the pruned filter.\\ 
These criteria impose significant constraints on the structure of the weights in $F$, but are necessary to ensure that each filter's feature representation remains consistent between their original task and any tasks for which they're shared during training. The most noteworthy constraint is that when storing the mask for $M_t$, all subnetworks that had been shared at the time of training $\Tau_t$ must also be included in the mask to satisfy the second constraint. We provide some additional discussion of this in the Supplementary Materials (SM).

\subsection{Measures of Usefulness}
When evaluating the properties of a subnetwork we use the following measures. For each measure we first collect all activations of the trainable layers in the full network $F$, computing usefulness scores as follows:\\
\textbf{Activation Score}: The activation score $A_{t',t}$ is calculated as the mean activation on task $\Tau_t$ over a given subnetwork $S_{t'<t}$. The mask $M_{t'}$ is applied to the collected activations and used to calculate $A_{t',t}$ as follows:
\begin{equation}\label{Def: Acts}
    A_{t',t} = \frac{1}{N_tL}\sum\limits_{n = 1} ^{N_t}\sum\limits_{l = 1}^{L}\big(\frac{1}{|I_{t
    '}|}\sum\limits_{i \in I^l_{t'}}f^l_i(X_{n,t})\big),
\end{equation}
Where $I^l_{t'}$ are the indices of all filters in layer $l$ included in $S_{t'}$. The activations are averaged first over the $I^l_{t'}$ filters in each given layer $l$, and then over the $N_t$ samples in $\Tau_t$ and the $L$ layers in $S_{t'}$.


\textbf{Connectivity}: The total connectivity score $P_{t',t}$ for a subnetwork $S_{t'<t}$ on task $\Tau_t$ is defined as as:
\begin{equation}\label{Def: Delta}
  P_{t',t}:=  \frac{1}{L}\sum\limits_{l=1}^{L}\rho^l_{t',t}, \;\;\;\hbox{where}
\end{equation}
\begin{equation}\label{Def: LayerRho}
  \rho^l_{t',t}:= \frac{1}{|I^{l}_{t'}|| J^{l+1}_{t'}|}\sum\limits_{i\in I^l_{t'}}\sum\limits_{j\in J^{l+1}_{t'}} \rho(f^{l}_{i,t'},f^{l+1}_{j,t'}|C_t).
\end{equation}
Here $I^l_{t'}$ and $J^{l+1}_{t'}$ denote the indices of filters in connected layers $l$ and $l+1$ which are included in $S_{t'}$,  
and $\rho(f^{l}_{i,t'},f^{l+1}_{j,t'})$ indicates the Pearson correlations between the activations of a pair of filters in those layers, conditioned on the classes of task $\Tau_t$. Only filters included in $S_{t'}$ are included in the calculation for $P_{t',t}$.\\
For the experiments in this paper we additionally square the values of $f^l_i(X_{n,t})$ for $A_{t',t}$ and $\rho(f^{l}_{i,t'},f^{l+1}_{j,t'})$ for $P_{t',t}$. This is intended to help visualize differences between dissimilar tasks and to make the usefulness measure strictly positive. We report the effect of this processing in the SM.

\begin{figure}[t]
    \begin{subfigure}[b]{\columnwidth}
    \centering
        \includegraphics[width=0.75\columnwidth]{"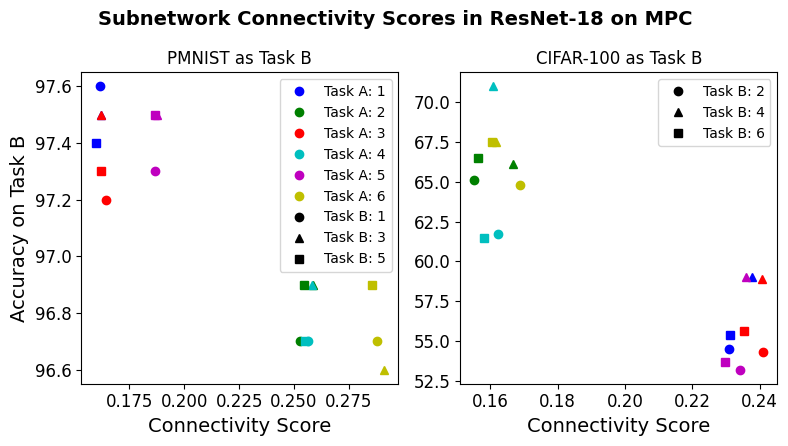"}
    \end{subfigure}
    \begin{subfigure}[b]{\columnwidth}
    \centering
        \includegraphics[width=0.75\columnwidth]{"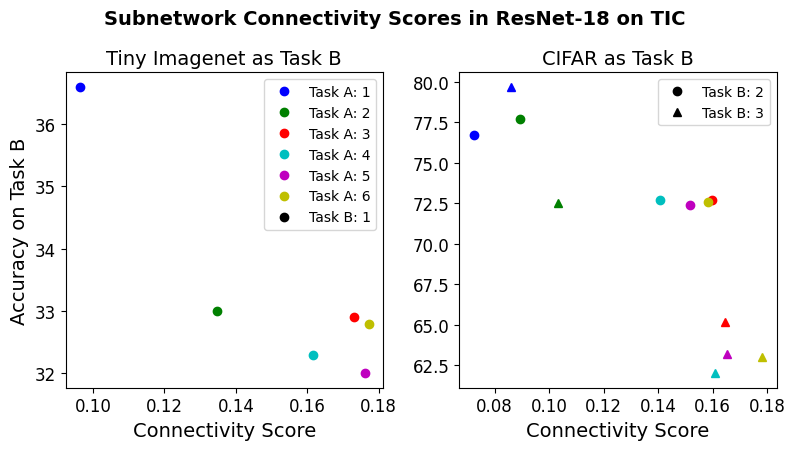"}
    \end{subfigure}
\caption{The mean subnetwork connectivities across all layers are plotted for each potential choice of $\Tau_\alpha$. For tasks in MPC (top), and TIC (bottom) we see that higher accuracy is strongly associates with a lower connectivity score $P$.}
\label{Fig:TwoTaskScatter}
\end{figure}

\textbf{Forward Knowledge Transfer}: The FKT between two tasks $\Tau_{t'}$ and a later task $\Tau_{t>t'}$ is defined as the difference in accuracy (Acc) between an individual network $F$ trained on $\Tau_t$ and the accuracy when $S_{t'}$ is shared prior to training $F_{+t'}$ on $\Tau_t$, and is calculated as:
\begin{equation}
  FKT_{t',t}:=  Acc(F_{+t'}(\mathbf{X}_t),Y_t)-Acc(F(\mathbf{X}_t),Y_t).
\end{equation}
In practice we don't compare directly to $F$, but between the networks $F_{+t'}$ resulting from different sharing decisions. By finding the task $\Tau_{t'}$ which maximizes $FKT_{t',t}$ for a given task $\Tau_t$, we seek to identify which tasks are most beneficial to share weights between.

\section{Experiments}
In order to investigate the impact that task similarity and complexity have on the usefulness of sharing a given subnetwork in facilitating FKT, we run three sets of experiments. These experiments are designed to first identify how our measures can predict the which weights are useful to share, and then demonstrate the impact of omitting weights based on their predicted usefulness, and finally to observe how the overall accuracy is impacted by applying weight sharing during the training of a dataset of six sequential tasks following the calculated subnetwork usefulness scores.

{\bf Setup} For experiments we perform 60 epochs of training, prune 65\% of non-frozen filters per layer, and do 20 epochs of finetuning after pruning. Training is done with an SGD optimizer at a learning rate of 0.1 which follows a scheduled decay by a factors of 10 down to $10^{-4}$. For compatibility, MNIST images are resized to 3x32x32, while Tiny Imagenet remains at its original 3x64x64 size. We use both VGG-16 and ResNet-18 models. To make the methods compatible with the subnetwork constraints, batch normalization layers are run without learned parameters or running statistics, and the residual connections of ResNet-18 are replaced by 1x1 convolutional layers in order to enable masking residual connections in the structured pruning setting. Details of model changes are given in the SM. For all experiments, the results are reported as the mean over three trials.

\begin{algorithm}[t]
    \SetAlgoLined 
    Given a dataset composed of tasks $\Tau_{1:T}$\\
    Assign $\Tau_\alpha$ and $\Tau_\beta$ as any two tasks in the dataset\\
    Initialize an untrained network $F$\\
    Denote $E_{tr}$ and $E_{ft}$ as the number of epochs for training and finetuning\\
     \For{$e = $ 1, \dots, $E_{tr}$}
     {
         Train $F$ on task $\Tau_\alpha$
     }
    Prune $k\%$ of filters from each layer in $F$\\
    Freeze weights and store binary mask $M_\alpha$\\
    \For{$e = $ 1, \dots, $E_{ft}$}
     {
         Train $S_\alpha$ on task $\Tau_\alpha$
     }
     Reinitialize pruned weights and share $S_\alpha$ for $\Tau_\beta$\\
    \For{$e = $ 1, \dots, $E_{tr}$}
     {
         Train $F_{+\alpha}$ on task $\Tau_\beta$
     }
     Evaluate test accuracy $Acc^\alpha_{\beta}$ of $F_{+\alpha}$ on task $\Tau_\beta$\\
     Evaluate usefulness measures $A_{\alpha,\beta}$ and $P_{\alpha,\beta}$ of $S_\alpha$ for $\Tau_\beta$\\
    Report $Acc^\alpha_{\beta}$, $A_{\alpha,\beta}$, $P_{\alpha,\beta}$
    \caption{Two Task Subnetwork Sharing}
    \label{Alg1-2task}
\end{algorithm}

\begin{algorithm}[t]
    \SetAlgoLined 
    Given a dataset composed of tasks $\Tau_{1:T}$\\
    Assign $\Tau_\alpha$, $\Tau_\beta$, and $\Tau_\gamma$ as three different tasks in the dataset\\
    Initialize an untrained network $F$\\
    Denote $E_{tr}$, $E_{ft}$ the epochs for training and finetuning\\
    \For{$i$ in $[\alpha, \beta, \gamma]$}
     {
         Reinitialize any pruned weights\\
         \If {$i = \beta$} 
         {
            Omit $S_\alpha$ for $\Tau_\beta$
        }
         \ElseIf {$i = \gamma$}
         {
         Choose to omit $S_\alpha$ and/or $S_\beta$
         }         
        \For{$e = $ 1, \dots, $E_{tr}$}
        {
             Train $F$ on task $\Tau_i$. 
        }
        Prune $k\%$ of unfrozen filters from each layer in $F$\\
        Freeze weights and store binary mask $M_i$\\
        \For{$e = $ 1, \dots, $E_{ft}$}
        {
             Train $S_i$ on task $\Tau_i$
        }
     }
    Report the test accuracy of $F_{+\alpha\beta}$, $F_{-\beta}$, or $F_{-\alpha}$ as  $Acc^{\alpha,\beta}_\gamma$,  $Acc^{\alpha}_\gamma$, or  $Acc^{\beta}_\gamma$, respectively.
    \caption{Three Task Subnetwork Omission}
    \label{Alg2-3task}
\end{algorithm}

{\bf Datasets} We utilize three different datasets, each a sequence of 6 tasks, to investigate the mechanism of FKT. The first dataset, Tiny Imagenet-CIFAR (\textbf{TIC}) is a variation of split CIFAR-10/100 (Golkar et al., 2019) in which task \textbf{1} is Tiny Imagenet (Le and Yang, 2015), task \textbf{2} is CIFAR-10, and tasks \textbf{3}-\textbf{6} are 10-class subsets of CIFAR-100 (Krizhevsky et al., 2009). We add Tiny Imagenet to better evaluate the impact of task complexity. The second dataset, Mixed PMNIST/CIFAR (\textbf{MPC}) has permuted copies of the MNIST (PMNIST) dataset as is commonly used in the permuted MNIST benchmark dataset for CL for tasks \textbf{1},\textbf{3}, and \textbf{5}. Subsets of CIFAR-100 are used for tasks \textbf{2},\textbf{4}, and \textbf{6}. This dataset is similar to the mixed dataset constructed in (Wang et al., 2020) and emphasizes the impact of task similarity on FKT. Lastly, Mixed KEF-MNIST/CIFAR (\textbf{KEF}) keeps the alternating CIFAR-100 subsets, but replaces the three PMNIST tasks with KMNIST-49 (Clanuwat et al., 2018), EMNIST-balanced (Cohen et al., 2017), and Fashion-MNIST (Xiao et al., 2017) for tasks \textbf{1}, \textbf{3}, and \textbf{5} respectively. These tasks are intended to serve as more challenging alternatives to PMNIST which are still dissimilar to CIFAR. As KEF's mix of both task complexity and similarity provides less clearly interpretable results, the experiments in the main paper primarily focus on the simpler datasets with additional results on KEF being provided in the SM.
\subsection{Two-Task Knowledge Transfer}
The simplest approach to evaluating the usefulness of a subnetwork in facilitating FKT is to consider the training of two tasks. We denote these two tasks $\Tau_\alpha$ and $\Tau_\beta$, drawn from any of the six tasks in a given dataset. After training on $\Tau_\alpha$, $F$ is pruned to generate the subnetwork $S_\alpha$, which is shared during training of $\Tau_\beta$. We can then evaluate the usefulness of sharing $S_\alpha$ given different choices of tasks for $\Tau_\alpha$. This allows us to see which tasks provide better FKT to each other task, and compare the resulting accuracy to the usefulness score calculated for the subnetwork $S_\alpha$ on $\Tau_\beta$. We outline the steps of this approach in Algorithm \ref{Alg1-2task}. \\
For a task $\Tau_\beta$ we can evaluate FKT and determine the usefulness of sharing the knowledge from different tasks by comparing the relative resulting accuracies for different choices of $\Tau_\alpha$. We compare this against the measures $A_{\alpha,\beta}$ and $P_{\alpha,\beta}$ for the shared subnetwork $S_\alpha$. A representative example of this comparison is shown in Figure \ref{Fig:TwoTask}. We observe that activations and connectivities tend to decrease in the subsequent layers, while the connectivity tends to be significantly lower for subnetworks $S_\alpha$ which yield better accuracy $Acc^\alpha_{\beta}$. We provide a clearer demonstration of this effect in Figure \ref{Fig:TwoTaskScatter}. This demonstrates that the lower connectivities aren't tied to a given task. In the MPC dataset CIFAR tasks have lower connectivity and yield the best accuracy when task $\Tau_\beta$ is also CIFAR, however when $\Tau_\beta$ is any PMNIST task then $S_\alpha$ from CIFAR tasks provide worse accuracy and higher connectivities. Comparing the plots for TIC, we see that more complex tasks such as CIFAR-10 and Tiny Imagenet provide a higher accuracy and lower connectivities when selected as $\Tau_\alpha$. In the SM we provide further examples of these results across each network and dataset.

\begin{table*}[t]
     \caption{\textbf{Impact of Subnetwork Omission on Task $\Tau_\gamma$ Accuracy}: The tables show how accuracy differs for ResNet-18 on the MPC dataset (left) and VGG-16 on the TIC dataset (right) for different sharing decisions on a set of potential 3-task sequences. We report accuracy on the third task $\Tau_\gamma$ when neither subnetwork was omitted ($Acc^{\alpha\beta}_\gamma$), or subnetworks $S_\beta$ or $S_\alpha$ were omitted ($Acc^{\alpha}_\gamma$ and $Acc^{\beta}_\gamma$ respectively). The highest accuracy for each sequence is shown in bold. For MPC, as task $\Tau_1$ is PMNIST if $\Tau_\beta$ and $\Tau_\gamma$ are both CIFAR ($\Tau_2$,$\Tau_4$,$\Tau_6$), omitting $S_\alpha$ may lead to slightly improved accuracy while omitting $S_\beta$ leads to significantly worse accuracy. For TIC where task $Tau_1$ is Tiny Imagenet, we often see the highest accuracy when omitting $S_\beta$, supporting that sharing only the most useful weights can outperform sharing all weights.}
     \label{tab:3taskomit}
    \begin{subtable}[h]{0.5\textwidth}
        \centering
        \begin{tabular}[t]{@{}llllccc@{}}
        \toprule
         Dataset & $\Tau_\alpha$ & $\Tau_\beta$ & $\Tau_\gamma$ & $Acc^{\alpha\beta}_{\gamma}$ & $Acc^{\alpha}_{\gamma}$ & $Acc^{\beta}_{\gamma}$   \\
        \midrule
         \multirow{20}{*}{\shortstack[l]{MPC}} 
         &\multirow{20}{*}{\shortstack[l]{1}} 
         &  \multirow{4}{*}{\shortstack[l]{2}}
            &   3   & 96.8 & \textbf{97.1} & 97.0 \\
             \cline{4-7}
          & & &       4   & \textbf{64.5} & 57.4 & 63.9 \\
             \cline{4-7}
          & & &       5   & 96.7 & \textbf{97.1} & 96.8 \\
             \cline{4-7}
          & & &       6   & 62.8 & 51.9 & \textbf{63.4} \\
             \cline{3-7}

         & & \multirow{4}{*}{\shortstack[l]{3}}
            &       2   & \textbf{50.9} & 50.4 & 49.9 \\
             \cline{4-7}
          & & &       4   & 56.0 & \textbf{57.9} & 56.8 \\
             \cline{4-7}
          & & &       5   & 97.3 & 97.0 & \textbf{97.4} \\
             \cline{4-7}
          & & &       6   & 51.9 & 51.2 & \textbf{52.4} \\
             \cline{3-7}

         &  &\multirow{4}{*}{\shortstack[l]{4}}
            &       2   & \textbf{59.2} & 49.2 & 59.1 \\
             \cline{4-7}
          & & &       3   & 96.8 & \textbf{97.1} & 96.9 \\
             \cline{4-7}
          & & &       5   & 96.8 & \textbf{97.1} & 96.7 \\
             \cline{4-7}
          & & &       6   & 60.8 & 51.5 & \textbf{61.0} \\
             \cline{3-7}

         &  &\multirow{4}{*}{\shortstack[l]{5}}
            &       2   & 50.0 & 49.0 & \textbf{50.8} \\
             \cline{4-7}
          & & &       3   & 97.2 & 97.1 & \textbf{97.3} \\
             \cline{4-7}
          & & &       4   & 57.0 & 56.6 & \textbf{57.1} \\
             \cline{4-7}
          & & &       6   & \textbf{51.7} & 50.4 & 51.3 \\
             \cline{3-7}
         & & \multirow{4}{*}{\shortstack[l]{6}}
            &       2   & \textbf{61.2} & 50.3 & 59.4 \\
             \cline{4-7}
          & & &       3   & 97.0 & \textbf{97.2} & 96.9 \\
             \cline{4-7}
          & & &       4   & 63.9 & 58.0 & \textbf{64.1} \\
             \cline{4-7}
          & & &       5   & 96.8 & \textbf{97.1} & 96.7 \\
        \bottomrule
    \end{tabular}    
    \end{subtable}
    \hfill
    \begin{subtable}[h]{0.5\textwidth}
        \centering
        \begin{tabular}[t]{@{}llllccc@{}}
        \toprule
         Dataset & $\Tau_\alpha$ & $\Tau_\beta$ & $\Tau_\gamma$ & $Acc^{\alpha\beta}_{\gamma}$ & $Acc^{\alpha}_{\gamma}$ & $Acc^{\beta}_{\gamma}$   \\
        \midrule
        \multirow{20}{*}{\shortstack[l]{TIC}}
        &\multirow{20}{*}{\shortstack[l]{1}} 
         &  \multirow{4}{*}{\shortstack[l]{2}}
            &       3   & 76.4 & \textbf{77.0} & 75.3 \\
             \cline{4-7}
          & & &       4    & 73.6 & 73.0 & \textbf{74.0} \\
             \cline{4-7}
          & & &       5   & \textbf{79.1} & 78.7 & 78.1 \\
             \cline{4-7}
          & & &       6    & 73.2 & 72.7 & \textbf{74.4} \\
             \cline{3-7}

         & & \multirow{4}{*}{\shortstack[l]{3}}
            &       2    & \textbf{81.2} & 80.6 & 81.1 \\
             \cline{4-7}
          & & &       4    & 64.6 & \textbf{74.2} & 64.2 \\
             \cline{4-7}
          & & &       5   & 73.27 & \textbf{78.5} & 73.0 \\
             \cline{4-7}
          & & &       6   & 66.8 & \textbf{72.6} & 66.6 \\
             \cline{3-7}

         &  &\multirow{4}{*}{\shortstack[l]{4}}
            &       2    & 80.5 & \textbf{80.7} & \textbf{80.7} \\
             \cline{4-7}
          & & &       3   & 69.8 & \textbf{77.9} & 66.1 \\
             \cline{4-7}
          & & &       5   & 70.9 & \textbf{78.7} & 70.7 \\
             \cline{4-7}
          & & &       6   & 65.6 & \textbf{72.5} & 65.7 \\
             \cline{3-7}

         & & \multirow{4}{*}{\shortstack[l]{5}}
            &       2   & 80.5 & \textbf{80.6} & \textbf{80.6} \\
             \cline{4-7}
          & & &       3   & 68.9 & \textbf{77.2} & 68.5 \\
             \cline{4-7}
          & & &       4   & 63.6 & \textbf{74.1} & 63.7 \\
             \cline{4-7}
          & & &       6    & 66.2 & \textbf{72.5} & 66.5 \\
             \cline{3-7}
         & & \multirow{4}{*}{\shortstack[l]{6}}
            &       2    & 80.5 & \textbf{81.0} & 80.5 \\
             \cline{4-7}
          & & &       3    & 68.9 & \textbf{77.6} & 68.1 \\
             \cline{4-7}
          & & &       4   & 62.9 & \textbf{73.2} & 64.2 \\
             \cline{4-7}
          & & &       5   & 71.7 & \textbf{78.1} & 71.7 \\
        \bottomrule
    \end{tabular}    
     \end{subtable}
\end{table*}

\subsection{Three-Task Subnetwork Omission}
While the previous experiment demonstrates which tasks improve FKT when shared, we also need to observe the effects of omitting certain subnetworks. It's been shown that simply sharing all frozen weights can provide sub-optimal accuracy, so here we look at how well the appropriate decision for omitting an entire subnetwork aligns with the expected usefulness of the available subnetworks following our findings in the previous experiment. For this we instead train networks on sequences of three tasks, $\Tau_\alpha$, $\Tau_\beta$, and $\Tau_\gamma$ chosen from among the tasks in a given dataset. \\
Following Algorithm \ref{Alg2-3task}, we train $F$ such that $S_\alpha$ is not shared during the training of $S_\beta$. This ensures that $S_\beta$ can be shared without $S_\alpha$, without impacting the representations of the features detected by filters in $S_\beta$. Before training $S_\gamma$, either $S_\beta$ or $S_\alpha$ are manually chosen to be omitted, or both $S_\alpha$ and $S_\beta$ are shared for $\Tau_\gamma$. After the training on $\Tau_\gamma$, we calculate the accuracy on $\Tau_\gamma$ as $Acc^{\alpha\beta}_\gamma$, $Acc^{\alpha}_\gamma$, $Acc^{\beta}_\gamma$, depending on whether both $S_\alpha$ and $S_\beta$ were shared, $S_\beta$ was omitted, or $S_\alpha$ was omitted, respectively. A subset of results are provided in Table \ref{tab:3taskomit}, however due to the amount of possible sequences the rest are located in the SM. \\
These results support the findings from the two-task experiments, and we can see that for the MPC dataset if both $\Tau_\beta$ and $\Tau_\gamma$ are CIFAR tasks ($\Tau_2, \Tau_4, \Tau_6)$ then omitting $S_\beta$ is significantly detrimental while omitting $S_\alpha$, which is a PMNIST task, can occasionally improve accuracy on $\Tau_\gamma$. This reflects the previous results that sharing similar tasks yields higher accuracies. Meanwhile results on TIC demonstrate that for most cases when $\Tau_\gamma$ is a CIFAR-100 task ($\Tau_{3-6}$), the best accuracy is achieved by omitting $S_\beta$ and relying only on the shared subnetwork $S_\alpha$ trained on the more complex Tiny Imagenet dataset. This pattern is less evident when CIFAR-10 is chosen for $\Tau_{\beta=2}$, likely because CIFAR-10 is of a more comparable complexity to Tiny Imagenet than the smaller subsets of CIFAR-100 are.

\begin{table*}[t]
\caption{\textbf{Six-Task Subnetwork Sharing:} ResNet-18 was sequentially trained on MPC, TIC, or KEF datasets under different weight sharing strategies, with the final test accuracies for each task and final average overall accuracy being reported. The strategies denoted "Share" share the N subnetworks with the lowest calculated connectivity score. Strategies denoted "Omit" omit the subnetwork with the lowest score and share the rest. Manual strategies for MPC and KEF share MNIST tasks with each other and CIFAR tasks with each other, while for TIC they only share $S_1$. For all datasets the manual strategy provides the best results. For MPC and KEF, omitting all subnetworks or omitting the subnetwork with the lowest connectivity yields high accuracy as well. For TIC, omitting the subnetwork with the lowest score significantly reduces overall accuracy.}\label{tab:6taskcl}
\centering
    \setlength{\tabcolsep}{5pt} 
        \begin{tabular}[t]{@{}lllcccccccc@{}}
        \toprule
         Dataset & Share Type & Share Order & N Shared & $Acc_1$ & $Acc_2$ & $Acc_3$ & $Acc_4$ & $Acc_5$ & $Acc_6$ & Mean $Acc$  \\
        \midrule
         \multirow{5}{*}{\shortstack[l]{MPC}} 
            &       Share   & All  & 6 & 97.4 & 47.7 & 96.6 & 50.7 & 96.1 & 42.5 & 71.84 \\
             \cline{2-11}
            &       Share   & None  & 0 & 97.5 & 65 & 96.9 & 61.7 & 94.9 & 58.9 & 79.14 \\
             \cline{2-11}
            &       Share   & Lowest  & 1 & 97.6 & 45.2 & 96.4 & 50.8 & 95.8 & 42.4 & 71.35\\
             \cline{2-11}
            &       Omit    & Lowest  & $t-1$ & 97.5 & 63.9 & 95.9 & 57.2 & 95.4 & 59.2 & 78.18 \\
             \cline{2-11}
            &       Share    & Manual  & - & 97.5 & 65.1 & 96.6 & 63.9 & 96.0 & 59.6 & \textbf{79.79} \\
        \cline{1-11}
         \multirow{5}{*}{\shortstack[l]{TIC}} 
            &       Share   & All  & 6 & 38.9 & 74.9 & 78.3 & 72.7 & 77.6 & 73.4 & 69.29 \\
             \cline{2-11}
            &       Share   & None & 0  & 38.8 & 74.6 & 59.1 & 50.1 & 56.7 & 54.7 & 55.68 \\
             \cline{2-11}
            &       Share   & Lowest  & 1  & 38.6 & 74.8 & 78.1 & 72.3 & 77.3 & 72.8 & 68.97 \\
             \cline{2-11}
            &       Omit    & Lowest  & $t-1$  & 39.0 & 75.1 & 63.4 & 63.7 & 68.9 & 60.1 & 61.69 \\
             \cline{2-11}
            &       Share    & Manual  & -  & 39.1 & 75.1 & 78.3 & 72.0 & 78.3 & 73.9 & \textbf{69.45} \\
        \cline{1-11}
         \multirow{5}{*}{\shortstack[l]{KEF}} 
            &       Share   & All  & 6 & 89.4 & 53.2 & 94.0 & 59.9 & 90.4 & 49.6 & 72.7 \\
             \cline{2-11}
            &       Share   & None & 0  & 89.1 & 63.1 & 95.3 & 61.1 & 90.6 & 56.0 & 75.88 \\
             \cline{2-11}
            &       Share   & Lowest  & 1  & 89.2 & 52.6 & 93.4 & 59.8 & 89.8 & 51.5 & 72.68 \\
             \cline{2-11}
            &       Omit    & Lowest  & $t-1$  & 89.1 & 65.1 & 93.9 & 62.3 & 90.2 & 49.3 & 74.96 \\
             \cline{2-11}
            &       Share    & Manual  & -  & 89.2 & 64.4 & 93.8 & 61.6 & 90.4 & 59.3 & \textbf{76.45} \\            
    \bottomrule
    \end{tabular}    
\end{table*}

\subsection{Six-Task Subnetwork Sharing}
To apply the results of the previous experiments in a conventional CL setting, we sequentially train a network over all tasks in a given dataset. We establish a set of simple strategies which dictate the subnetwork sharing decisions based on the $P_{t',t}$ score for each subnetwork $S_{t'<t}$. These strategies calculate $P_{t',t}$ as detailed in \ref{Def: Delta}. For available subnetworks $S_{t'<t}$, we either share all subnetworks, share no subnetworks, share the subnetwork with the lowest calculated score, or omit the subnetwork with the lowest score and share the rest. This is done at the start of each task, and any subnetworks shared or omitted during task $\Tau_{t-1}$ are re-evaluated for task $\Tau_{t}$. For each dataset we include experiments where we have manually chosen which tasks to share based on our insights from the previous experiments. For MPC and KEF this means sharing MNIST tasks with only other MNIST tasks and CIFAR tasks with other CIFAR tasks, while for TIC it means sharing only $S_1$. The results of applying these strategies are provided in Table \ref{tab:6taskcl}. The strategy of sharing the subnetwork with the lowest connectivity performed poorly in MPC, while omitting the lowest subnetwork and sharing the rest performed significantly better. We see a similar pattern in KEF, which is composed of a similar pattern of MNIST and CIFAR tasks. For TIC we observe different results, with omitting the subnetwork with the lowest connectivity giving the worst results. For all datasets the manually crafted sharing strategies, which were informed by the insights from earlier experiments matched or exceeded the accuracy of the other strategies.

\section{Discussion}
CL makes an interesting and complex setting to investigate the mechanisms involved in training a DNN. In this work this is made evident by some of the differences we observe between the two-task and six-task experiments. In the former, we observe that the subnetworks $S_\alpha$ with the lowest connectivities yield a better accuracy for $F_{+\alpha}$ on $\Tau_\beta$. This coincided with two intuitive cases, the first being when $\Tau_\alpha$ and $\Tau_\beta$ are similar, as when they're both a version of PMNIST or a subset of CIFAR-100 in the MPC dataset or KEF dataset (shown in the SM). This makes sense as features useful for one MNIST or CIFAR tasks would be expected to be beneficial in other similar tasks. In the TIC dataset we see the second intuitive case for high FKT, which is for the more complex or informative tasks in the dataset. For TIC this is Tiny Imagenet and to a lesser extent CIFAR-10. While each subset of CIFAR-100 has 10 classes and 5,000 training images, CIFAR-10 has 50,000 training images and Tiny Imagenet has 100,000 64x64 training images spanning 200 classes. The idea that complex tasks are beneficial for transfer learning isn't novel, however our experiments clearly demonstrate this behavior in a controlled subnetwork setting for CL. \\
These insights were supported in the three-task experiments, where we choose between sharing one or both subnetworks $S_\alpha$ and/or $S_\beta$ when training a third task $\Tau_\gamma$. This showed that often oversharing can lead to detriments to accuracy as was seen in work by Wang et al. (2020). For instance, in cases where $\Tau_\beta$ is CIFAR and $\Tau_\gamma$ is PMNIST in MPC, we see that omitting the unrelated task $\Tau_{\beta}$ performs best. When all three tasks in MPC belong to the same type (e.g. for $\alpha,\beta,\gamma$ sequences 1,3,5 and 1,5,3) then the sharing decision affects the resulting accuracy significantly less as either $S_\alpha$ and $S_\beta$ alone are sufficient for FKT. For the cases where $\Tau_\alpha$ is Tiny Imagenet, as expected we see that sharing only $S_\alpha$ provides high accuracy for most sequences. Surprisingly comparing $Acc^{\alpha\beta}_\gamma$ and $Acc^\beta_\gamma$ reveals that for the former, even though $S_\alpha$ is being shared and has been structured such that the feature representation remains consistent in between $F_{+\alpha}$ and $F_{+\alpha\beta}$, the resulting accuracy is highly similar to the sub-optimal $Acc^{\beta}_\gamma$. These results also support the idea that oversharing can be of significant detriment to CL. \\
While these experiments strongly support connectivity-based sharing strategies, the six-task experiments demonstrate the complexity of CL problems. For MPC we see that rather than sharing the subnetwork with the lowest connectivity, we observe higher overall accuracies when we omit this subnetwork and share all other subnetworks or share no subnetworks. While this seems at odds with the suggested role of connectivity, identifying which subnetworks are shared for each tasks reveals the issue. \\
Due to the implemented subnetwork constraints detailed in Section \ref{section:interpretablepruning}, if we share a subnetwork $S_{t'<t}$ for task $\Tau_t$, that subnetwork must remain part of $M_t$ and is shared as a part of $S_t$ to ensure that the filters within $S_t$ behave similarly to when they were trained on task $\Tau_t$. This is why we observe that in this strategy of simply sharing the lowest-connectivity subnetwork, task $\Tau_2$ shares $S_1$ as it is the only available subnetwork. For each subsequent task, often the most recent subnetwork $S_{t-1}$ has the lowest connectivity for task $\Tau_t$. This creates a chain where mask $M_3$ includes $S_2$, $M_4$ includes $S_3$, etc, and by sharing a single subnetwork we're effectively sharing all past subnetworks in order to maintain the imposed structural constraints. We suspect that the reason for the most recent subnetworks having lower connectivity is that including subnetworks trained on different tasks within $S_t$ reduces the overall flow of information within the subnetwork. An alternative sharing behavior can be seen in the Omit strategy for TIC. Likely due to its complexity, Tiny Imagenet is routinely the subnetwork with the lowest connectivity score, and is omitted by every task. Not leveraging Tiny Imagenet significantly worsens overall accuracy compared to other strategies. We circumvent these issues of systematic sharing strategies in our manual experiments, which demonstrate the potential for sharing based on subnetwork connectivity or task similarity and complexity. 

\section{Conclusion}
Although the environment of controlled experiments demonstrates potential for the use of connectivity in dictating weight sharing decisions, the simple sharing strategies implemented in this work don't apply well to longer sequences of tasks. This is believed to be in part due to the structural constraints implemented on pruning and sharing for the sake of improving the interpretability of results. In a practical CL approach it may be infeasible to work around these constraints, and some shift in filter behavior between their original task and any subsequent task with which they're shared should be expected. There are multiple potential and compelling future directions which could leverage the findings from this paper. Based on prior CL work which had used connectivity in the pruning process (Andle and Yasaei Sekeh, 2022) we had expected higher values to coincide with better accuracies on $\Tau_\beta$. Given that connectivity is intended as a reflection of the information flow of the network, it may be that a lower connectivity is less suited for the original task, but is more generalizable for the new task. Similarly, while we see compelling evidence that sharing similar tasks is beneficial in the two-task and three-task experiments, it remains to be seen how only sharing subnetworks trained on similar tasks may impact the adversarial robustness or generalizability of the resulting network compared to subnetworks from more diverse tasks.


\section{Acknowledgment}
This work has been partially supported by NSF CDS\&E-MSS (2053480), Cisco, and Maine Space Grant Consortium (MSGC) seed grant; the findings are those of the authors only and do not represent any position of these funding bodies.

\renewcommand{\refname}{References}

\bibliographystyle{IEEEbib}

\nocite{*} 

\appendix
\section{Overview of Appendices}
To supplement the discussion and results provided in the main paper, we provide here a further discussion and explanation of details of the implemented methods. This discussion covers a more thorough explanation and justification for the implemented constraints on the subnetwork structure and sharing, a comparison of the effect of processing the usefulness measures ($A$ and $P$) through squaring them, and a more detailed description of the modifications made to the ResNet-18 model and why they were made. We additionally provide extended results for each experiment. These include providing plots of connectivity and activations against accuracy for all datasets and architectures used, additional task sequences for the three-task experiments shown in the main paper, and the results of the shown six-task experiment strategies on VGG16 for all datasets.                                                                    

\section{Additional Discussion and Details of Methods}
\subsection{Discussion of Subnetwork Constraints}
In Section \ref{section:interpretablepruning} we discuss the constraints which are imposed on the pruning and sharing of subnetworks. To briefly reiterate, those are:
\begin{enumerate}
    \item Once filter $f^l_i$ is frozen, its weights cannot be updated.
    \vspace{-0.2cm}
    \item If $f^{l}_i$ is shared, filters in previous layers with which it is connected by non-zero weights are shared as well.
    \vspace{-0.2cm}
    \item During pruning, when a filter $f^l_i$ is pruned, all weights in $F^{l+1}$ connected to $f^l_i$ are pruned. 
\end{enumerate}
The explanation of the rationale for these is best understood through a simple example. Say we have three tasks, $\Tau_\alpha$,$\Tau_\beta$,$\Tau_\gamma$, and have already trained subnetworks $S_\alpha$ and $S_\beta$ where $S_\alpha$ was shared for the training of $\Tau_\beta$. To train the network for task $\Tau_\gamma$ we may share one or both subnetworks as in the Three-Task experiments. The need for constraint 1 is simple and ubiquitous in architectural methods of Continual Learning (CL). Say we share $S_\alpha$ and then retrain the weights in it for $\Tau_\gamma$, which is a significantly different task than $\Tau_\alpha$. Because we use a single shared network and distinguish the filters used for each task through masking, if we were to now perform inference on task $\Tau_\alpha$ we would be using filters trained on $\Tau_\gamma$, leading to Catastrophic Forgetting (CF). Since $S_\alpha$ had been used when training $S_\beta$ then this would also cause CF to occur for $\Tau_\beta$ indirectly.\\
To demonstrate the need for constraint 2, consider that we share only the filters frozen during the training of $\Tau_\beta$, without sharing any of $S_\alpha$. Consider a filter $i$ in layer $F^l$, $f^l_{i,\beta}$ in $S_\beta$ which was originally optimized to take as input the weighted activations from layer $F^{l-1}$. Since $S_\alpha$ had been shared during this training, a subset of the layer $F^{l-1}$ deemed $F^{l-1}_\alpha$ have non-zero activations and contribute to the feature detected by $f^l_{i,\beta}$. In such a case, the feature which $f^l_{i,\beta}$ detects in the inputs is dependent on features in part provided by $F^{l-1}_\alpha$. By not sharing $S_\alpha$ alongside $S_\beta$ during $\Tau_\gamma$, $f^l_{i,\beta}$ will not have access to these features upon which it initially depended for its learned role and it is expected to identify features for $\Tau_\gamma$ which are not representative of its role in $\Tau_\beta$. If we wish to share weights solely for their usefulness on the current task this may not be an issue, but this difference in behavior limits claims that can be made about \textit{why} $f^l_{i,\beta}$, or $S_\beta$ as a whole, are useful when shared for $\Tau_\gamma$. As our aim is to compare which tasks are useful when shared with each other and why, we enforce constraint 2 to ensure that $S_\alpha$ is included as a subset of $S_\beta$ such that all dependencies are shared alongside each other.\\
For constraint 3 consider a filter $f^l_{i,\alpha}$ trained during $\Tau_\alpha$. During pruning of $S_\alpha$, a subset of filters in $F^{l-1}_\alpha$ have all of their weights set to zero. Any weights connecting these filters to $f^l_{i,\alpha}$ will have no impact on the value of $f^l_{i,\alpha}$, since the activations being weighted are all zero. However, if this subset of filters in $F^{l-1}_\alpha$ are trained and frozen as part of $S_\beta$, then sharing both tasks for $\Tau_\gamma$ results in the opposite problem of constraint 2. If those weights connecting $f^l_{i,\alpha}$ are non-zero, then the now non-zero activation values of the newly trained filters in $F^{l-1}_\alpha$ will influence the behavior of $f^l_{i,\alpha}$ compared to the optimized role it had for $\Tau_\alpha$. To avoid this unpredictable affect we can simply prune such hanging connections. Together constraints 2 and 3 ensure that we can share multiple tasks in a modular manner without unrelated subnetworks interfering with one another.\\
As noted, these constraints are restrictive on the structure of the subnetworks, with the most prominent issue being that constraint 2 can result in cases where every subsequent subnetwork trained shares the previous subnetwork, and by extension all other subnetworks, effectively voiding the benefits of modularity offered by the subnetwork approach in the first place. This is why we suggest in the Conclusion that, while we feel these constraints are necessary for this type of investigation, they are likely too overly restrictive for a real-world problem.

\subsection{Comparison of Squaring Usefulness Measures}
\begin{figure*}[t]
    \begin{subfigure}[b]{\textwidth}
    \centering
        \includegraphics[width=1.0\columnwidth]{"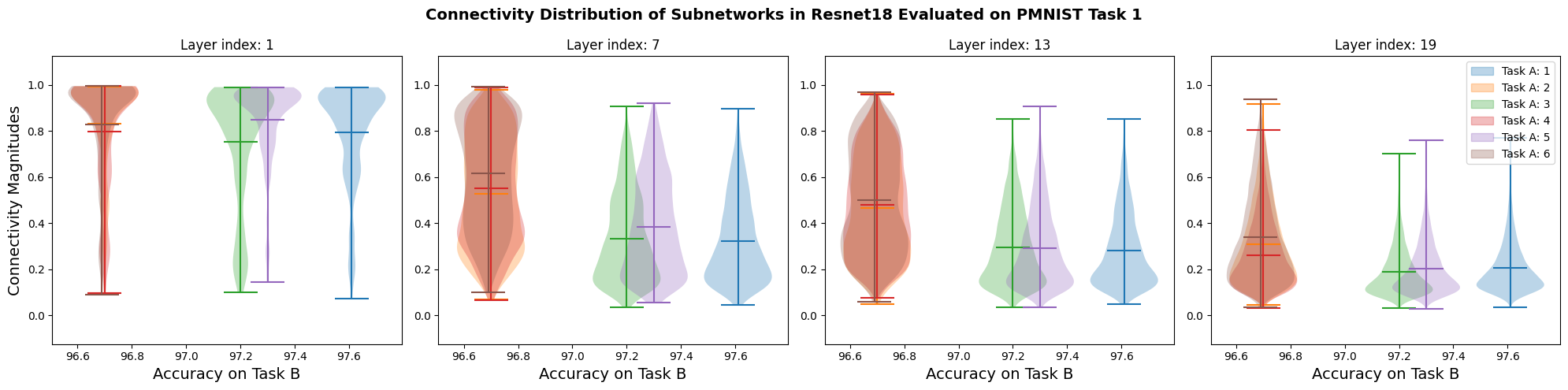"}
    \end{subfigure}
    \begin{subfigure}[b]{\textwidth}
    \centering
        \includegraphics[width=1.0\columnwidth]{"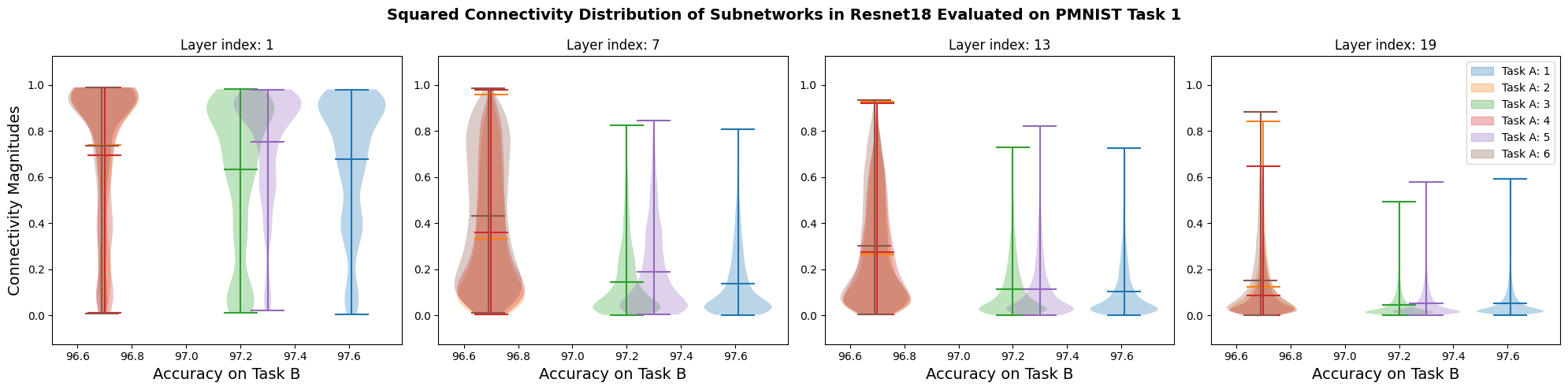"}
    \end{subfigure}
\caption{The connectivity distributions for ResNet-18 on a PMNIST task from MPC are plotted for four convolutional layers. The connectivity values are reported raw (top) or after being squared (bottom).}
\label{Fig:SquaringSM}
\end{figure*}

To predict the usefulness of a given subnetwork $S_{t'}$ on a later task when shared, we calculate either $A_{t',t}$ or $P_{t',t}$ (and by extension $\rho^l_{t',t}$). In practice we square the values of the activations when calculating $A_{t',t}$ and $\rho^l_{t',t}$ when calculating $P_{t',t}$. The reasoning for this is to more strongly discriminate between high and low values as well as to ensure that the measures are strictly positive. In particular the latter is done for two reasons, first is that we average these values both within a layer and over all layers, and want to avoid cases where negative values canceling out positive values within or across layers may obscure the trends in behavior. The second reason is that the initial hypothesis behind using measures derived from the activations on the current tasks data was to see the magnitude of the response of $S_{t'}$ to $\Tau_t$, regardless of the sign of those responses. It is worth noting that this initial reasoning is not fully supported by the findings that subnetworks with lower connectivity yield better accuracy when shared. As shown in Figure \ref{Fig:SquaringSM} we find that this processing doesn't interfere with the trends in connectivity, which is the measure of usefulness used throughout the experiments. Notably connectivity is defined as the absolute value of the correlation between a pair of filters' activations, so the benefits of maintaining strictly positive values aren't applicable in this case.

\subsection{Modifications to ResNet-18}
When implementing pruning-based CL methods, and in particular structured pruning as is done in this work, ResNets architecture poses a challenge due to the residual connections which result in a nonlinear set of paths through the network. Consider two layers $F^l$ and $F^{l+n}$ which have the same number of filters and are connected by a residual connection such that the outputs of $F^l$ are added back in to the outputs of $F^{l+n}$ prior to $F^{l+n+1}$. If a given filter $f^{l+n}_i$ is pruned but $f^{l}_i$ is not, then the nonzero outputs of $f^l_i$ will be added back into the outputs of the pruned filter $f^{l+n}_i$. This behavior results in pruned filters having non-zero activations, which interferes with the constraints we have implemented on weight pruning and sharing. Rather than constrain the pruning by forcing both layers to be pruned identically, we replace the residual connections which are normally implemented as element-wise addition functions with 1x1 convolutional layers, similar to the down-sampling layers. This allows us to store masks for the residual connections as in any other trainable layer, allowing layers $F^l$ and $F^{l+n}$ to have different pruned filter indices while mapping the outputs of $F^l$ to a set of residual activations compatible with the pruning of $F^{l+n}$.\\
In addition to this change, for both VGG-16 and ResNet-18 we implement batch normalization layers without affine parameters or running statistics. Although these parameters could be learned for each task and stored on a task-by-task basis, they need to then be reloaded whenever we perform inference on a given task. This increases the complexity and potentially leads to discrepancies of behavior of a shared filter between its original task and the current task if these parameters differ between the two tasks. To circumvent these concerns for the sake of simplicity and to maintain consistency we remove these parameters.

\section{Additional Experiment Results}
In order to supplement the representative subset of results shown in the main experiments we provide here the remaining results for each experiment. All results are the mean or aggregate results over three trials.

\subsection{Two-Task Experiments}
\begin{figure*}[t]
    \begin{subfigure}[b]{\textwidth}
    \centering
        \includegraphics[width=0.9\columnwidth]{"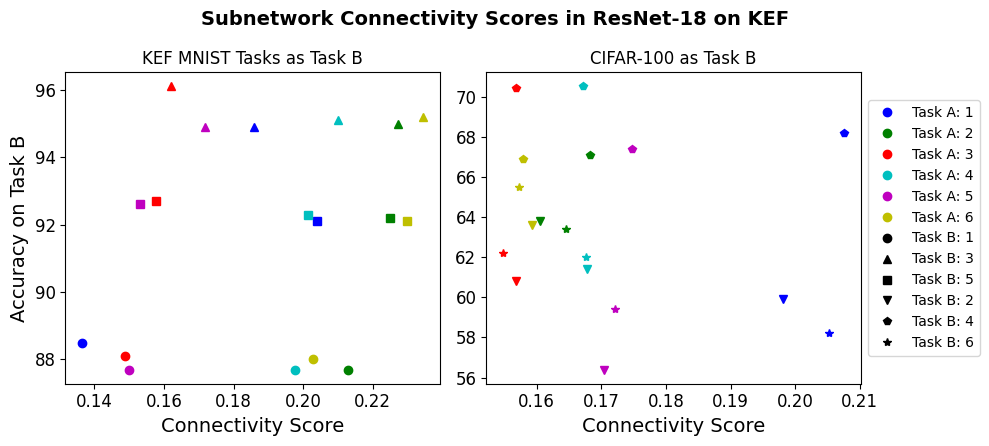"}
    \end{subfigure}
    \begin{subfigure}[b]{\textwidth}
    \centering
        \includegraphics[width=0.9\columnwidth]{"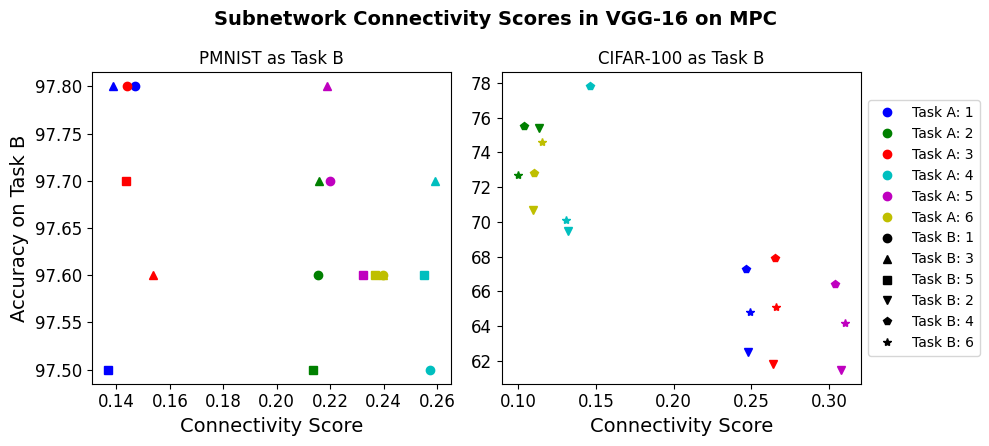"}
    \end{subfigure}
    \begin{subfigure}[b]{\textwidth}
    \centering
        \includegraphics[width=0.9\columnwidth]{"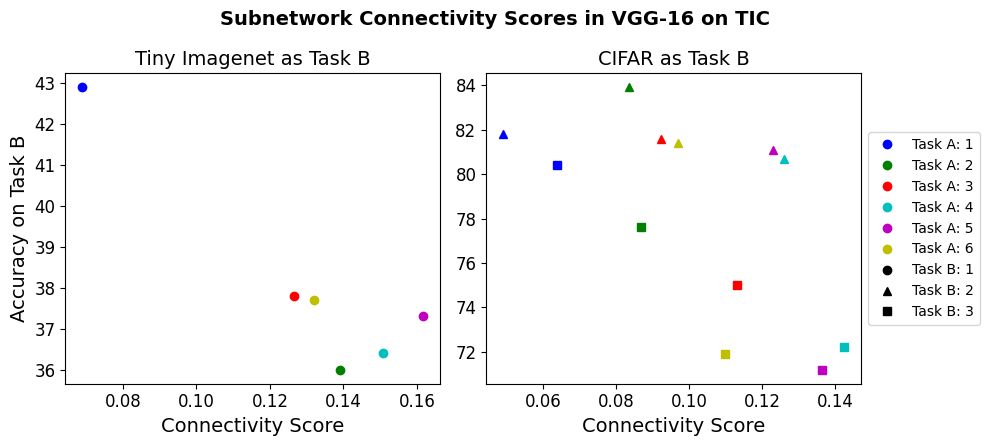"}
    \end{subfigure}
    \begin{subfigure}[b]{\textwidth}
    \centering
        \includegraphics[width=0.9\columnwidth]{"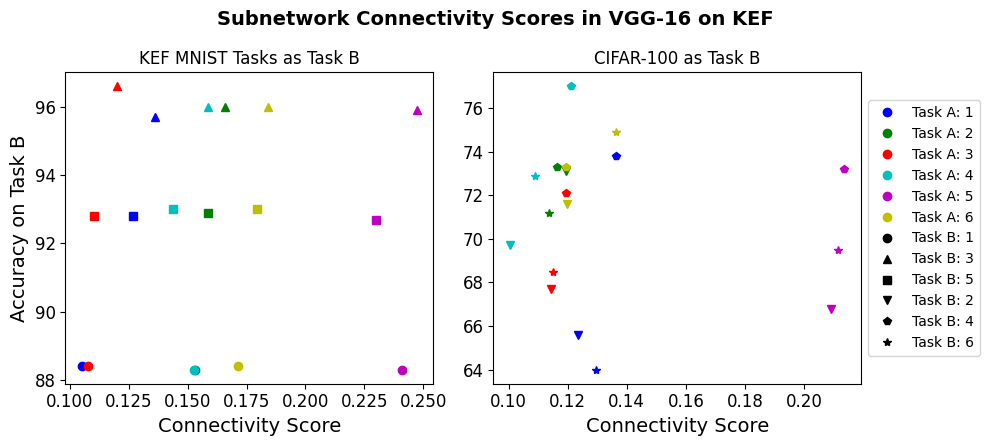"}
    \end{subfigure}    
\caption{The connectivity scores $P$ are plotted against accuracy for each remaining Two-Task experiment. These results demonstrate trends similar to those observed in the main experiments in which lower connectivity scores coincide with a higher resulting $Acc_\beta$. Among the datasets KEF shows the least clear trends, which may be due to the diversity of both task complexity as well as task similarity compared to the other two datasets which are designed to emphasize only one or the other.}
\label{Fig:TwoTaskSM}
\end{figure*}

The Two-Task experiments demonstrated the relationship between connectivity and activations within a subnetwork $S_\alpha$ when shared for a subsequent task $\Tau_\beta$. Figure \ref{Fig:TwoTaskSM} demonstrates that the trends observed in the main experiments extend to each other dataset and architecture used in this investigation. Of the implemented datasets, a relationship between connectivity and accuracy is hardest to observe in KEF. This is potentially due to the KEF being designed to vary among tasks both in task complexity and similarity, whereas TIC primarily varies in complexity (with Imagenet and CIFAR-10 being much more complex than CIFAR-100 subsets) or task similarity (with similar PMNIST and CIFAR tasks in MPC). This may result in relationships between tasks becoming less readily evident.

\subsection{Three-Task Experiments}
The Three-Task experiments trained two subnetworks $S_\alpha$ and $S_\beta$ (where $S_\alpha$ wasn't shared during $\Tau_\beta$) and then for a given task $\Tau_\gamma$ either shares one or both of these subnetworks. Given that there are six tasks in each dataset and we're selecting three, with the assumption that order matters since earlier tasks are allocated more filters and aren't affected by any prior tasks' pruning or freezing decisions, we only reported results over one choice of $\Tau_\alpha$ for two dataset/architecture combinations in the main experiments. Here we report additional results for those two combinations in Table \ref{tab:3taskomitSM}. For brevity and readability we opt not to report over all sequences, but as MPC is composed of 3 MNIST and 3 CIFAR-100 tasks and TIC tasks 3-6 are all CIFAR-100 subsets, the provided results are sufficiently representative of the remaining task orders. The reported results support the findings of the main experiments, as in MPC if tasks $\Tau_\alpha$ and $\Tau_\gamma$ are similar (such as when they're both PMNIST or both CIFAR-100) then omitting $\Tau_\beta$ often yields the best performance. For TIC, the results support that omitting the subnetwork trained on the less complex CIFAR-100 is generally optimal if the other subnetwork was trained on the more complex CIFAR-10 or Tiny Imagenet tasks.\\

\begin{table*}[t]
     \caption{\textbf{Impact of Subnetwork Omission on Task $\Tau_\gamma$ Accuracy}: The tables show additional results from the Three-Task experiments for ResNet-18 on the MPC dataset (left) and VGG-16 on the TIC dataset (right). We report accuracy on the third task $\Tau_\gamma$ when neither subnetwork was omitted ($Acc^{\alpha\beta}_\gamma$), or subnetworks $S_\beta$ or $S_\alpha$ were omitted ($Acc^{\alpha}_\gamma$ and $Acc^{\beta}_\gamma$ respectively). The highest accuracy for each sequence is shown in bold. The results support the findings of the main experiments, as for MPC when tasks $\Tau_\alpha$ and $\Tau_\gamma$ belong to the same type of task (PMNIST or CIFAR) sharing only $S_\alpha$ often outperforms other options. Similarly for TIC when $\Tau_{\alpha=2}$ is CIFAR-10, in most cases omitting $S_\beta$ outperforms sharing both subnetworks similar to as was observed when $\Tau_\alpha$ was Tiny Imagenet. When $\Tau_\alpha$ is CIFAR-100 and $\Tau_\beta$ is Tiny Imagenet or CIFAR-10 then omitting $S_\alpha$ frequently yields the best accuracy.}
     \label{tab:3taskomitSM}
    \begin{subtable}[h]{0.5\textwidth}
        \centering
        \begin{tabular}[t]{@{}llllccc@{}}
        \toprule
         Dataset & $\Tau_\alpha$ & $\Tau_\beta$ & $\Tau_\gamma$ & $Acc^{\alpha\beta}_{\gamma}$ & $Acc^{\alpha}_{\gamma}$ & $Acc^{\beta}_{\gamma}$   \\
        \midrule
         \multirow{40}{*}{\shortstack[l]{MPC}} 
         &\multirow{20}{*}{\shortstack[l]{2}} 
         &  \multirow{4}{*}{\shortstack[l]{1}}
            &   3   &     \textbf{97.4}     &     96.8     &     97.3     \\
             \cline{4-7}
          & & &       4  &    57.0      &    \textbf{65.0 }     &    56.6      \\
             \cline{4-7}
          & & &       5  &     \textbf{97.5 }    &    96.8      &    97.3      \\
             \cline{4-7}
          & & &       6  &    52.1      &    \textbf{64.7}      &   52.5       \\
             \cline{3-7}

         & & \multirow{4}{*}{\shortstack[l]{3}}
            &       1  &    \textbf{97.3}      &   96.4       &   \textbf{97.3 }      \\
             \cline{4-7}
          & & &       4 &   57.6       &  \textbf{ 65.9}       &   56.8       \\
             \cline{4-7}
          & & &       5 &   97.2       &    96.5      &  \textbf{ 97.3 }      \\
             \cline{4-7}
          & & &       6 &  51.1        &    \textbf{65.3}      &   52.2       \\
             \cline{3-7}

         &  &\multirow{4}{*}{\shortstack[l]{4}}
            &       1 &   96.8       &    96.5      &   \textbf{96.9 }      \\
             \cline{4-7}
          & & &       3 &   \textbf{97.0  }     &    96.8      &   \textbf{97.0 }      \\
             \cline{4-7}
          & & &       5  &   \textbf{96.9 }      &    96.6      &   96.8       \\
             \cline{4-7}
          & & &       6 &    62.6      &    \textbf{64.8}      &   63.1       \\
             \cline{3-7}

         &  &\multirow{4}{*}{\shortstack[l]{5}}
            &       1  &   97.2       &     96.4     &   \textbf{ 97.3  }    \\
             \cline{4-7}
          & & &       3  &  \textbf{ 97.3 }      &    96.7      &  \textbf{ 97.3  }     \\
             \cline{4-7}
          & & &       4 &  57.3        &    \textbf{ 65.6  }   &    57.9      \\
             \cline{4-7}
          & & &       6  &    53.0      &   \textbf{ 64.1 }     &   53.8       \\
             \cline{3-7}
         & & \multirow{4}{*}{\shortstack[l]{6}}
            &       1  &    \textbf{96.9  }    &     96.4     &  \textbf{ 96.9  }     \\
             \cline{4-7}
          & & &       3  & \textbf{  97.0 }      &    96.8      &   96.9       \\
             \cline{4-7}
          & & &       4 &   64.1       &  \textbf{  66.7 }     &    64.5      \\
             \cline{4-7}
          & & &       5 &   \textbf{96.7  }     &    96.4      &   \textbf{ 96.7 }     \\

        \cline{2-7}
        &\multirow{20}{*}{\shortstack[l]{3}} 
         &  \multirow{4}{*}{\shortstack[l]{1}}
            &   2        &   \textbf{ 51.9 }     &    49.2      &   51.4       \\
             \cline{4-7}
          & & &       4  &   \textbf{ 59.1 }     &     57.6     &   58.1       \\
             \cline{4-7}
          & & &       5  &   \textbf{ 97.5 }     &   97.0       &   97.4       \\
             \cline{4-7}
          & & &       6 &   \textbf{53.4 }      &     50.1     &   52.3       \\
             \cline{3-7}

         & & \multirow{4}{*}{\shortstack[l]{2}}
            &       1 &    96.8      &    \textbf{97.1 }     &    96.9      \\
             \cline{4-7}
          & & &       4 &   64.2       &    56.1      &   \textbf{65.8 }      \\
             \cline{4-7}
          & & &       5  &   96.8       &    \textbf{97.1}      &   96.9       \\
             \cline{4-7}
          & & &       6 &  63.6        &    48.6      & \textbf{65.2 }        \\
             \cline{3-7}

         &  &\multirow{4}{*}{\shortstack[l]{4}}
            &       1  &   96.8       &   \textbf{97.1  }     &   96.9       \\
             \cline{4-7}
          & & &       2 &   \textbf{61.1 }      &     50.5     &  59.8        \\
             \cline{4-7}
          & & &       5  &   96.7       &   \textbf{97.1 }      &   96.9       \\
             \cline{4-7}
          & & &       6  & \textbf{ 60.7 }       &    50.7      &   58.5       \\
             \cline{3-7}

         &  &\multirow{4}{*}{\shortstack[l]{5}}
            &       1  &   \textbf{97.3}       &     97.2     & 97.1         \\
             \cline{4-7}
          & & &       2 &  50.3        &   48.1       &   \textbf{51.8}       \\
             \cline{4-7}
          & & &       4  &  \textbf{ 58.2  }     &      57.8    &  57.0        \\
             \cline{4-7}
          & & &       6  & \textbf{51.1  }       &   50.2       &   50.9       \\
             \cline{3-7}
         & & \multirow{4}{*}{\shortstack[l]{6}}
            &       1  &   96.8       & \textbf{ 97.3 }       &      96.9    \\
             \cline{4-7}
          & & &       2 &     \textbf{61.7}     &     48.1     &    61.3      \\
             \cline{4-7}
          & & &       4  &   \textbf{66.4 }      &    57.5      &   64.7       \\
             \cline{4-7}
          & & &       5 &   96.7       &  \textbf{ 97.0 }      &    96.9      \\
        \bottomrule
    \end{tabular}    
    \end{subtable}
    \hfill
    \begin{subtable}[h]{0.5\textwidth}
        \centering
        \begin{tabular}[t]{@{}llllccc@{}}
        \toprule
         Dataset & $\Tau_\alpha$ & $\Tau_\beta$ & $\Tau_\gamma$ & $Acc^{\alpha\beta}_{\gamma}$ & $Acc^{\alpha}_{\gamma}$ & $Acc^{\beta}_{\gamma}$   \\
        \midrule

         \multirow{40}{*}{\shortstack[l]{TIC}} 
         &\multirow{20}{*}{\shortstack[l]{2}} 
         &  \multirow{4}{*}{\shortstack[l]{1}}
            &   3       &     77.9     &     76.1     &     \textbf{79.4 }    \\
             \cline{4-7}
          & & &       4  &    74.2      &     \textbf{ 76.0}    &   73.2       \\
             \cline{4-7}
          & & &       5  &    78.2      &    \textbf{ 79.0}     &   78.5       \\
             \cline{4-7}
          & & &       6  &     \textbf{74.9 }    &  73.2        &    74.0      \\
             \cline{3-7}

         & & \multirow{4}{*}{\shortstack[l]{3}}
            &       1  &  \textbf{35.1 }       &    33.6      &  34.7        \\
             \cline{4-7}
          & & &       4 &   64.2       &  \textbf{75.0 }       &   62.5       \\
             \cline{4-7}
          & & &       5 &  72.3        & \textbf{78.7 }        &   73.2       \\
             \cline{4-7}
          & & &       6 &  66.8        &    \textbf{73.1 }     &   66.5       \\
             \cline{3-7}

         &  &\multirow{4}{*}{\shortstack[l]{4}}
            &       1    &   33.9       &    32.8      &     \textbf{34.4}     \\
             \cline{4-7}
          & & &       3  &     69.7     & \textbf{ 76.5 }       &   69.5       \\
             \cline{4-7}
          & & &       5  &   70.8       &   \textbf{78.4}       &   71.7       \\
             \cline{4-7}
          & & &       6  &   64.9       &   \textbf{ 71.7}      &   65.1       \\
             \cline{3-7}

         &  &\multirow{4}{*}{\shortstack[l]{5}}
            &       1    &  \textbf{34.3 }       &    32.8      &    34.0      \\
             \cline{4-7}
          & & &       3  &    69.5      &   \textbf{76.5 }      &   68.3       \\
             \cline{4-7}
          & & &       4  &   65.4       &   \textbf{ 72.2 }     &   65.4       \\
             \cline{4-7}
          & & &       6  &   65.4       &    \textbf{73.4}      &   64.7       \\
             \cline{3-7}
         & & \multirow{4}{*}{\shortstack[l]{6}}
            &       1    &   \textbf{ 35.5 }     &    32.8      &    34.9      \\
             \cline{4-7}
          & & &       3  &    70.0      &    \textbf{74.5 }     &   70.3       \\
             \cline{4-7}
          & & &       4  &   64.4       &   \textbf{ 76.5 }     &   64.6       \\
             \cline{4-7}
          & & &       5  &    71.9      &  \textbf{ 78.2 }      &   73.3       \\

        \cline{2-7}
        &\multirow{20}{*}{\shortstack[l]{3}} 
         &  \multirow{4}{*}{\shortstack[l]{1}}
            &   2        &    80.6      &    80.1      &    \textbf{ 81.2 }    \\
             \cline{4-7}
          & & &       4  &    68.0      &    62.1      &   \textbf{75.7 }      \\
             \cline{4-7}
          & & &       5  &    76.3      &   73.2       &  \textbf{79.6 }       \\
             \cline{4-7}
          & & &       6  &    71.4      &  67.7        &  \textbf{74.7 }       \\
             \cline{3-7}

         & & \multirow{4}{*}{\shortstack[l]{2}}
            &       1    &   34.4       &  \textbf{34.7 }       &   33.8       \\
             \cline{4-7}
          & & &       4  &   69.6       &  62.7        &  \textbf{74.9 }       \\
             \cline{4-7}
          & & &       5  &    73.0      &   73.0       & \textbf{ 77.2 }       \\
             \cline{4-7}
          & & &       6  &  69.9        &  65.8        &  \textbf{72.8 }       \\
             \cline{3-7}

         &  &\multirow{4}{*}{\shortstack[l]{4}}
            &       1    &   34.0       &   34.1       &   \textbf{34.2 }      \\
             \cline{4-7}
          & & &       2  &  \textbf{ 80.6 }      &    80.5      &  80.5        \\
             \cline{4-7}
          & & &       5  &   \textbf{71.5}       &    \textbf{ 71.5}     &    70.8      \\
             \cline{4-7}
          & & &       6  &  66.0        &   \textbf{ 67.9 }     &  66.2        \\
             \cline{3-7}

         &  &\multirow{4}{*}{\shortstack[l]{5}}
            &       1    &   \textbf{ 35.2 }     &    34.6      &   34.9       \\
             \cline{4-7}
          & & &       2  &   80.7       &   80.7       & \textbf{ 80.8}        \\
             \cline{4-7}
          & & &       4  &    \textbf{66.2 }     &   63.4       &   65.1       \\
             \cline{4-7}
          & & &       6  &   \textbf{67.2  }     &    67.1      & \textbf{ 67.2 }       \\
             \cline{3-7}
         & & \multirow{4}{*}{\shortstack[l]{6}}
            &       1    &   \textbf{34.9 }      &   34.6       &    \textbf{34.9 }     \\
             \cline{4-7}
          & & &       2  &   80.6       &   80.6       &   \textbf{81.0}       \\
             \cline{4-7}
          & & &       4  &   \textbf{66.0 }      &    62.6      &  63.8        \\
             \cline{4-7}
          & & &       5  &   72.0       &    72.3      &   \textbf{72.6}       \\
        \bottomrule
    \end{tabular}    
     \end{subtable}
\end{table*}

\subsection{Six-Task Experiments}
The Six-Task experiments demonstrate the overall accuracy on all tasks in a given dataset when implementing various subnetwork sharing strategies. To  supplement the results in the main experiments which report on ResNet-18 for each dataset, here we provide the results of VGG-16 on each dataset for the same sharing strategies. These results are shown in Table \ref{tab:6taskclSM}. Results on VGG-16 demonstrate similar behaviors as were seen for ResNet-18, however while the manual sharing strategies perform well relative to most strategies, they failed to outperform sharing all subnetworks for TIC or no subnetworks for MPC and KEF. This is expected to be due to differences between VGG-16 and ResNet-18, perhaps VGG-16 is more negatively affected by the structural constraints and therefore benefits more from sharing all or no subnetworks.

\begin{table*}[t]
\caption{\textbf{Six-Task Subnetwork Sharing:} VGG-16 was sequentially trained on a given dataset (MPC, TIC, or KEF) under different weight sharing strategies, with the final test accuracies for each task and final average overall accuracy being reported. The strategies are identical to those used in the main experiments for ResNet-18. Unlike with ResNet-18 although the manual sharing strategies outperformed many of the other alternatives, they still were unable to outperform sharing no subnetworks for MPC and KEF. Similar to with ResNet-18, one of the best strategies for TIC was to share all subsets as they're all similar and this ensures sharing of the subnetworks from Tiny Imagenet and CIFAR-10 with the other tasks.}\label{tab:6taskclSM}
\centering
    \setlength{\tabcolsep}{5pt} 
        \begin{tabular}[t]{@{}lllcccccccc@{}}
        \toprule
         Dataset & Share Type & Share Order & N Shared & $Acc_1$ & $Acc_2$ & $Acc_3$ & $Acc_4$ & $Acc_5$ & $Acc_6$ & Mean $Acc$  \\
        \midrule
         \multirow{5}{*}{\shortstack[l]{MPC}} 
            &       Share   & All  & 6 & 97.6 & 28.1 & 49.2 & 24.6 & 48.9 & 24.2 & 45.4 \\
             \cline{2-11}
            &       Share   & None  & 0 & 97.6 & 72.7 & 96.6 & 62.3 & 94.9 & 57.2 & \textbf{80.2} \\
             \cline{2-11}
            &       Share   & Lowest  & 1 & 97.6 & 35.7 & 96.6 & 30.1 & 44.1 & 25.3 & 54.9 \\
             \cline{2-11}
            &       Omit    & Lowest  & $t-1$ & 97.6 & 71.9 & 96.8 & 41.7 & 84.3 & 39.1 & 71.9 \\
             \cline{2-11}
            &       Share    & Manual  & - & 95.8 & 70.4 & 96.8 & 64.6 & 95.8 & 40.5 & 77.3 \\
        \cline{1-11}
         \multirow{5}{*}{\shortstack[l]{TIC}} 
            &       Share   & All  & 6 & 43.3 & 79.8 & 75.6 & 70.6 & 72 & 67.3 & \textbf{68.1} \\
             \cline{2-11}
            &       Share   & None & 0  & 43.4 & 81.6 & 65.3 & 51.1 & 46.2 & 48.8 & 56.1 \\
             \cline{2-11}
            &       Share   & Lowest  & 1  & 43.4 & 80.3 & 73.3 & 68.2 & 74.1 & 66.5 & 67.6 \\
             \cline{2-11}
            &       Omit    & Lowest  & $t-1$  & 43 & 81.3 & 75.3 & 59.4 & 73.6 & 52 & 64.1 \\
             \cline{2-11}
            &       Share    & Manual  & -  & 43.1 & 80.0 & 72.9 & 66.4 & 73.7 & 65.4 & 66.9 \\
        \cline{1-11}
         \multirow{5}{*}{\shortstack[l]{KEF}} 
            &       Share   & All  & 6 & 89.1 & 55.7 & 94.1 & 47.7 & 91.1 & 46.6 & 70.7 \\
             \cline{2-11}
            &       Share   & None & 0  & 89.0 & 73.0 & 95.8 & 62.7 & 89.5 & 53.5 & \textbf{77.2}\\
             \cline{2-11}
            &       Share   & Lowest  & 1 & 89.1 & 57.1 & 94.6 & 47.4 & 90.8 & 37.1 & 69.4 \\
             \cline{2-11}
            &       Omit    & Lowest  & $t-1$ & 88.9 & 73.6 & 95.1 & 54.7 & 91.5 & 55.2 & 76.5 \\
             \cline{2-11}
            &       Share    & Manual  & - & 89.1 & 71.8 & 94.7 & 55.3 & 90.9 & 53.1 & 75.8 \\
    \bottomrule
    \end{tabular}    
\end{table*}

\end{document}